\documentclass[letterpaper, 10 pt, conference]{ieeeconf}  

\IEEEoverridecommandlockouts                              
\usepackage{amsmath} 
\usepackage{amssymb}  
\usepackage{graphicx}
\usepackage{booktabs}
\usepackage{multirow}
\usepackage{stfloats}
\usepackage[T1]{fontenc}

\title{\LARGE \bf
Efficient Strategy Learning by Decoupling Searching and Pathfinding for Object Navigation
}

\author{Yanwei Zheng, \ Shaopu Feng, \ Bowen Huang, \ Chuanlin Lan, \ Xiao Zhang, \ Dongxiao Yu*
\thanks{Yanwei Zheng, Shaopu Feng, Bowen Huang,Chuanlin Lan, Xiao Zhang, Dongxiao Yu are with  Shandong University, Qingdao 266237, China. 
}
\thanks{
E-mail: {\tt\{zhengyw@sdu.edu.cn, fengsp@mail.sdu.edu.cn, huangbw@mail.sdu.edu.cn, lancl@sdu.edu.cn, xiaozhang@sdu.edu.cn, dxyu@sdu.edu.cn\}}. Dongxiao Yu is corresponding author (*).}
}

\begin{document}
\bibliographystyle{IEEEtran}

\maketitle
\thispagestyle{empty}
\pagestyle{empty}

\begin{abstract}

Inspired by human-like behaviors for navigation: first searching to explore unknown areas before discovering the target, and then the pathfinding of moving towards the discovered target, recent studies design parallel submodules to achieve different functions in the searching and pathfinding stages, while ignoring the differences in reward signals between the two stages. As a result, these models often cannot be fully trained or are overfitting on training scenes. Another bottleneck that restricts agents from learning two-stage strategies is spatial perception ability, since the studies used generic visual encoders without considering the depth information of navigation scenes. To release the potential of the model on strategy learning, we propose the Two-Stage Reward Mechanism (TSRM) for object navigation that decouples the searching and pathfinding behaviours in an episode, enabling the agent to explore larger area in searching stage and seek the optimal path in pathfinding stage. Also, we propose a pretraining method Depth Enhanced Masked Autoencoders (DE-MAE) that enables agent to determine explored and unexplored areas during the searching stage, locate target object and plan paths during the pathfinding stage more accurately. In addition, we propose a new metric of Searching Success weighted by Searching Path Length (SSSPL) that assesses agent's searching ability and  exploring efficiency. Finally, we evaluated our method on AI2-Thor and RoboTHOR extensively and demonstrated it can outperform the state-of-the-art (SOTA) methods in both the success rate and the navigation efficiency.
\end{abstract}

\section{Introduction}
Object navigation\cite{zhu2017target,duvtnet,mayo2021visual,dang2022unbiased} requires an agent to seek and move to the vicinity of the target object in an unfamilar environment, leveraging panoramic or egocentric views. Benefiting from the development of artificial intelligence, end-to-end methods\cite{mayo2021visual,zhang2021hi,pal2021learning} based on reinforcement learning\cite{mnih2016asynchronous,schulman2017proximal} are widely used in navigation tasks owing to their flexibility and adaptability, with the expectation to approach and even surpass human expert navigation performance.
\begin{figure}[htb]
    \centering
    \includegraphics[scale=0.43]{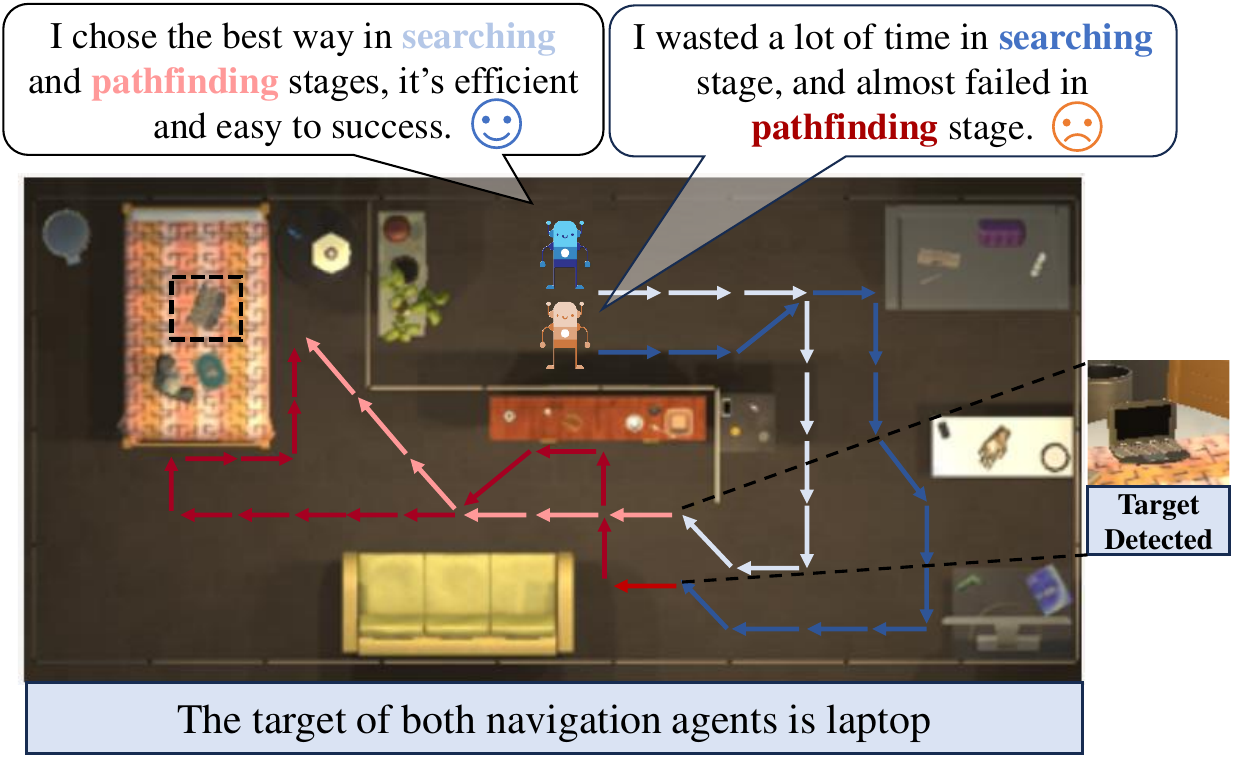}
    \caption{The core idea of our method. Existing methods have proposed the concept of two-stage navigation for agents, but the sparse and improper reward signals make it difficult for agents to select an efficient and successful one from countless paths and actions. Our method utilizes a two-stage reinforcement learning reward mechanism and perception enhancement of spatial depth information to guide agents to learn efficient exploration and navigation strategies, improving success rates.}
    \label{TSRM_fig}
\end{figure}
Since humans often exhibit different motivations and strategies before and after discovering the target, which inspires the studies of navigation agent. 

Prior studies have focused on developing efficient strategies for various stages by enhancing network structures. These enhancements have included the creation of modules tailored for specific purposes, such as searching based on object correlation \cite{du2020learning,dang2022unbiased,gadre2022continuous,lingelbach2023task} and pathfinding utilizing target object features\cite{dang2022search,IOM}.
While these improvements in model structure have led to enhanced performance, the increasing complexity of network architectures has posed new challenges. Specifically, the traditional sparse rewards used in training have exacerbated the difficulty of model convergence\cite{zhu2017target}. Furthermore, relying solely on rewards based on the shortest path can result in overfitting to the training scenarios.
In essence, the adoption of single-stage reward mechanisms overlooks the causal relationships between strategies employed at different stages, thereby limiting the model's ability to adapt to diverse situations.
Another critical bottleneck in enabling agents to learn effective strategies lies in inadequate spatial perception, particularly with regard to depth information. This limitation stems from the use of generic visual encoders, rather than using encoders specifically designed for navigation tasks, as the front-end of the model. Without this capability, agents are unable to discern which areas they have explored or unexplored, assess the distance to discovered targets, or plan viable paths to these targets. Consequently, a significant challenge remains in equipping agents with the ability to extract spatial depth information from monocular RGB images.
Lastly, despite various metrics have been proposed, there is still short of practical indicators  for evaluating the exploration efficiency of the agents during the searching stage.

\begin{figure*}[ht]
\centering
\includegraphics[scale=0.5]{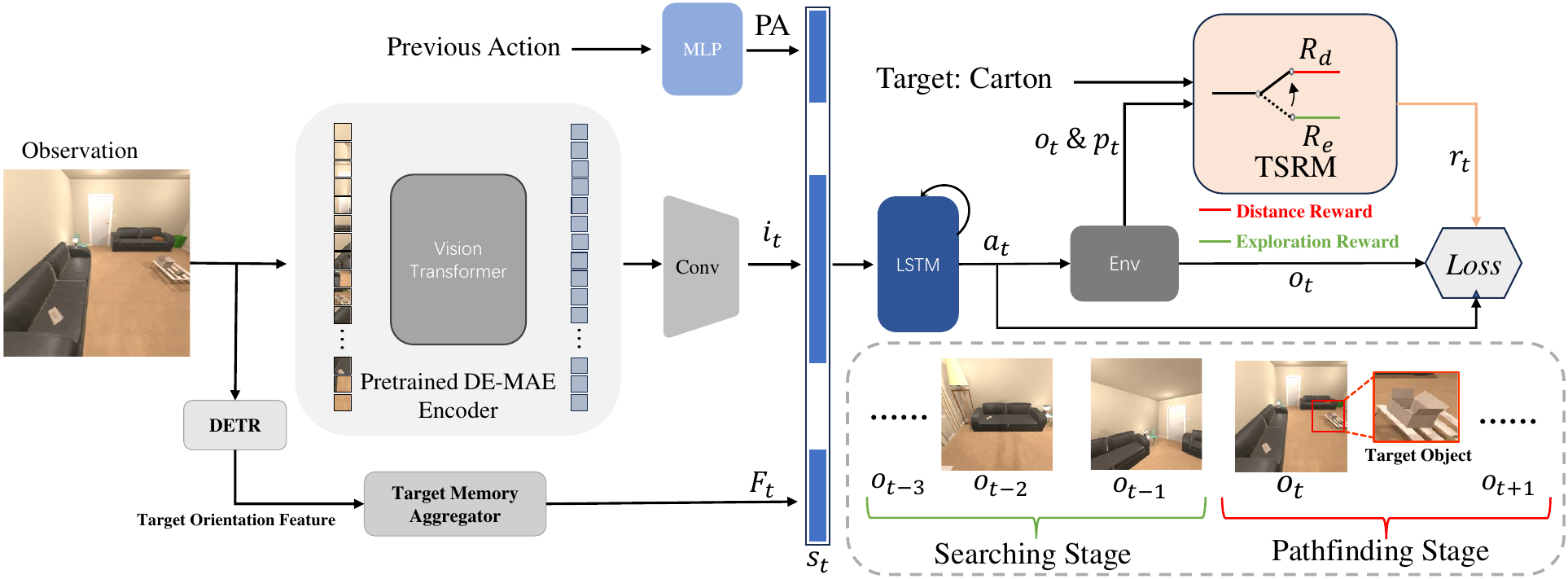}
\caption{Model overview. The agent obtains RGB image \begin{math} \boldsymbol{o}_{\boldsymbol{t}} \end{math} and target orientation feature processed by DETR. 
Vision transformer pretrained by DE-MAE serves as the image feature extractor followed by a convolutional layer and outputs the image embedding feauture \begin{math} \boldsymbol{i}_{\boldsymbol{t}} \end{math}  based on \begin{math} \boldsymbol{o}_{\boldsymbol{t}} \end{math}. 
The target memory aggregator proposed by IOM\cite{IOM} outputs target orientation embedding feature \begin{math} \boldsymbol{F}_{\boldsymbol{t}} \end{math}. \begin{math} \boldsymbol{i}_{\boldsymbol{t}} \end{math}, \begin{math} \boldsymbol{F}_{\boldsymbol{t}} \end{math} and previous action embedding \begin{math}
PA\end{math} are concatenated as the state representation \begin{math} \boldsymbol{s}_{\boldsymbol{t}} \end{math}. 
Finally, the LSTM policy network outputs the action distribution \begin{math} \boldsymbol{a}_{\boldsymbol{t}} \end{math} according to \begin{math} \boldsymbol{s}_{\boldsymbol{t}} \end{math}. 
The TSRM module determines the stage mode and the reward or punishment that should be given based on the historical and current location  \begin{math} \boldsymbol{p}_{\boldsymbol{t}} \end{math} and observation \begin{math} \boldsymbol{o}_{\boldsymbol{t}} \end{math}.
When the target object is first detected, the episode irreversibly enters the pathfinding stage from the beginning searching stage, then distance reward will be given to agent replacing exploration reward.}
\label{overview}
\end{figure*}

To release the potential of the models in strategy learning, this paper introduces a Two-Stage Reward Mechanism (TSRM) tailored for object navigation. During training, each episode is divided into the initial  searching stage and the subsequent pathfinding stage. The boundary between the stages is determined by the first observation frame of the target, achieved with a predefined level of confidence.
In searching stage, the reward is proportional to the area of the newly observed region, encouraging the agent to explore more effectively. In the pathfinding stage, the mechanism not only rewards or penalizes movement actions based on the agent's distance to the target but also takes into account the impact of rotation actions on reaching the target. This consideration is particularly crucial for agents with limited perspectives, which have often been overlooked in previous approaches.
Furthermore, to equip agents with a sharp "eye" for exploration and pathfinding, we propose a novel pretraining method tailored for navigation image encoders: Depth Enhanced Masked AutoEncoder (DE-MAE). . In this method, the RGB channels of the images serve as the input with certain patches randomly masked. Different from the classic  Masked AutoEncoder (MAE)\cite{he2022masked}, the decoder is designed to reconstruct both the RGB and depth information. Compared with previous pretraining algorithms, the encoder pretrained by DE-MAE excels at capturing the spatial information within images, meeting the requirements of indoor scene navigation tasks. 
Lastly, we introduce a new evaluation metric, Searching Success weighted by Searching Path Length (SSSPL) that assesses agent's searching and exploring efficiency by comparing its performance to the theoretically optimal path of searching stage, compensating for the lack of comparability in previous indicators\cite{anderson2018evaluation,dang2023multiple}.
The overall structure of our model is shown in Figure \ref{overview}. And our contributions can be summarized as follows:

\textbf{1)} We proposed the Two-Stage Reward Mechanism (TSRM), which encourages navigation agents to efficiently explore the unknown areas before discovering the target, and then to accurately locate the target and find feasible path to reach it.

\textbf{2)} We propose Depth Enhanced Masked AutoEncoder (DE-MAE), a self-supervised algorithms to train the image feature extractor for the navigation agent.

\textbf{3)} A new evaluation metric named Searching Success weighted by Searching Path Length (SSSPL) is proposed to evaluate the exploring efficiency in searching stage.

\section{Related Works}
\noindent \textbf{End-to-end navigation.}
End-to-end methods based on reinforcement learning  have gained prominence in navigation tasks due to their flexibility across diverse  environment. Initial investigations employed  convolutional neural network to encode image observation, coupling them with recurrent neural networks to serve as the agent's memory mechanism\cite{zhu2017target}. As research progressed, researchers have realized the instructive role of the correlation between the positions of different objects in navigation decision-making, such as using graph neural networks to extract the distribution between different types of target objects\cite{du2020learning,li2021ion,gadre2022continuous}, and constructing hierarchical relationships from objects, regions to scenes\cite{zhang2021hi}.
Furthermore, transformer \cite{vaswani2017attention} networks have seen widespread application in visual navigation tasks. They have been utilized both as feature extractors for images \cite{duvtnet, mayo2021visual} and as replacements for RNNs, capitalizing on their strengths in long-term memory and reasoning capabilities \cite{fukushima2022object, fang2019scene, chen2021history}.
Drawing inspiration from human navigation strategies, recent studies \cite{dang2022search, IOM} introduced the navigation thinking network. This innovative approach selectively retains high-confidence features of target objects, derived from DETR \cite{carion2020end}, to aid the agent in pinpointing the target object's location. 

\noindent \textbf{Self-supervised learning in visual navigation.}
Self-supervised learning has demonstrated its capacity to extract valuable knowledge from extensive unlabeled datasets\cite{he2020momentum,he2022masked,devlin2018bert,radford2018improving}.
In the domain of visual navigation, where images captured from the environment often lack annotations such as classification labels or textual descriptions, the application of self-supervised learning techniques for pretraining the image embedding extractor of navigation agents has gained considerable attention in recent years.  For instance, contrastive learning approaches has been employed to enhance navigation performance under conditions of sparse rewards \cite{yadav2023offline, majumdar2022ssl}, Similarly, OVRL-V2 \cite{yadav2023ovrl} utilizes the Masked Autoencoder (MAE) for pretraining purposes. Furthermore, VC-1 \cite{majumdar2024we} also leverages MAE and has built a substantial dataset tailored for universal embodied intelligence tasks, encompassing navigation, to train the image encoder.

\section{Methods}
\subsection{Problem Setting}\label{setting}
The agent is initialized to a random reachable position in the scene with random pitch and yaw angles: \begin{math}s=(x,y,\theta,\beta)\end{math}. A random target $g$ is assigned. According to the RGB image $o_t$ and target $g$, the agent learns a navigation strategy $\pi(a_t|o_t,g)$, where \begin{math}a_t\in A = \{MoveAhead, RotateLeft, \end{math} \begin{math}RotateRight, LookDown, LookUp, Done\} \end{math}. The output of $Done$ means the end of episode. 
Ultimately, When the agent is within
 1 meter of the target object and meanwhile target object appears in the agent's view, if $Done$ is output the episode is considered successful.
If the agent outputs  $Done$ without reaching the target or the total number of output actions exceeds the predefined maximum episode length, the navigation episode is failed.
\subsection{Two Stage Reinforcement Learning}\label{TSRL}
\begin{figure}[htb]
    \centering
    \includegraphics[scale=0.28]{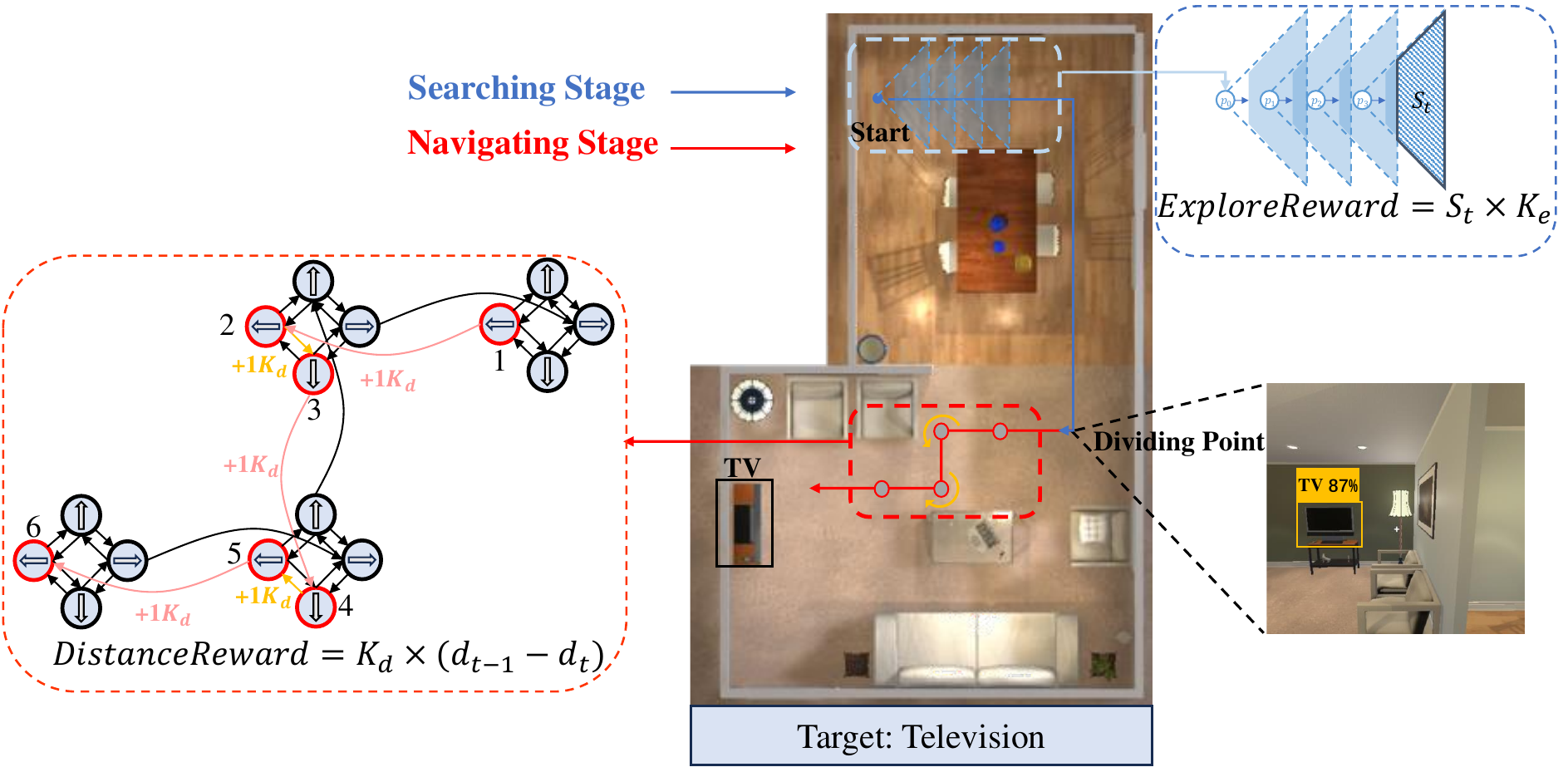}
    \caption{Two-Stage Reward Mechanism. In the first stage, the exploration reward  is proportional to the increase of the searched area which is the union of multiple trapezoids. After the target object is discovered, the navigation stage is entered. In this stage, each $Movehead$(pink) or $Rotation$(yellow) action will be rewarded or punished based on its impact to the distance between the agent and the target.}
    \label{TSRM_fig}
\end{figure}
When humans performs an object navigation task in an unknown room, they usually move around to explore at first and during exploration check whether the target object appears in the current view. If it does not appear, they continue searching for unexplored area. To enhance search efficiency, repeatedly searching the same area should be avoided. Once the target has been clearly observed, there is no need to continue searching. Instead, the focus shifts to locating the orientation of the target object and moving towards it. The two-stage learning approach for object navigation we propose aims to achieve this by employing a segmented design of the reward function, enabling the agent to learn the behavioral pattern of initially searching and subsequently finding a path to the target.
\paragraph{Searching Stage}
An episode must begin with searching stage. In this stage, we set the exploration reward to be proportional to the increase of the overall clearly observed area. It is difficult to define the exact size of this area especially in different environments and observation equipment, so we propose an estimation approach. From the bird's-eye view, we simplify the area that can be observed clearly at each reachable position into an isosceles trapezoid intercepted from a right triangle (as shown in Figure \ref{TSRM_fig}).

When the agent reaches a new position, its searched area increases by an isosceles trapezoid. By performing geometric union operations on these trapezoids $T_{0,1...t}$, the resulting polygon corresponds to the area that the agent has searched. Since the polygon may include some areas out of the room, we perform a intersection operation on the polygon and the room's overhead projection. The final polygon is searched region $RG_t$. The size of the exploration reward is directly related to the area difference between $RG_{t+1}$ generated after the agent takes a new action and $RG_t$ in the previous step.
$RG_0$ is initially null. Actually, we calculate searched region $RG_t$ more efficiently recursively, and the specific formulas for calculating searched region and  exploration reward are as follows:
\begin{align}
RG_{t}=&(RG_{t-1}\cup T_{t})\cup RoomBounds \\
ExploreReward_{t}=& K_e \times Area(RG_{t}-RG_{t-1})  \label{exreward}
\end{align}
where $K_e$ is a positve constant, $RoomBounds$ is the planar projection of the room, obtained from  AI2-Thor's API, and $Area(\cdot)$ is a function that calculates the area of a plane figure.
\paragraph{Pathfinding Stage}
If the target object can be clearly observed within current field of view, there is no need to search for unknown areas. Instead, the focus of navigation should shift to locating the target object and finding a feasible path to reach it. Previous navigation reward mechanisms have only considered changes in the Euclidean or geodesic distance from the agent's position to the target object, neglecting the impact of rotation actions that alter the agent's orientation. This is particularly crucial for agents without panoramic observation ability, as executing appropriate rotation actions at the right time is essential for subsequent movement. Conversely, making a wrong rotation action will not only cause the subsequent movement trajectory to deviate from the correct direction, but may also cause the target to disappear from the view.

As described in Fig \ref{TSRM_fig}, We process the scene as a grid map composed of passable position points, and convert the grid into a directed graph. Each directed edge in the graph corresponds to a $Moveahead$ or $Rotation$ action. The weights of all edges are $1$.
On this directed graph, the distance reward following the execution of an action is proportional to the reduction in the minimum distance to any potential successful terminal states . If this distance increases unexpectedly, a penalty is applied. The calculation formula for distance reward (penalty) is as follows:
\begin{align}
DistanceReward_{t}=K_d  \times (d(p_{t-1},g)-d(p_{t},g))\label{navreward}    
\end{align}

where $d(p,g)$ is the minimum distance from position $p$ to any terminal states that successfully reach the target $g$ within 1 meter, and $K_d$ is a positive constant.            
To improve computational efficiency, we use multi-source optimal path algorithm to preprocess the training scene grid data, thus avoid calculating the $d(p,g)$ during reinforcement training.
\paragraph{Dividing Point of Two Stages}
Entering pathfinding stage too early can cause the model to overfit in the training scenes, underestimating the importance of exploration; conversely, an excessively long searching stage will reduce navigation efficiency and the probability of discovering targets.

On dividing point, the agent should objectively see the target object for the first time, and be aware of this subjectively. A threshold value $C_{target}$ is set between 0 and 1. If in the detection results of DETR, the confidence of the corresponding target object in the current frame is greater than $C_{target}$ for the first time, the agent enters the pathfinding stage from searching stage, the reward calculation mechanism also changes.
\subsection{Depth Enhanced MAE}
\begin{figure}
    \centering
    \includegraphics[scale=0.29]{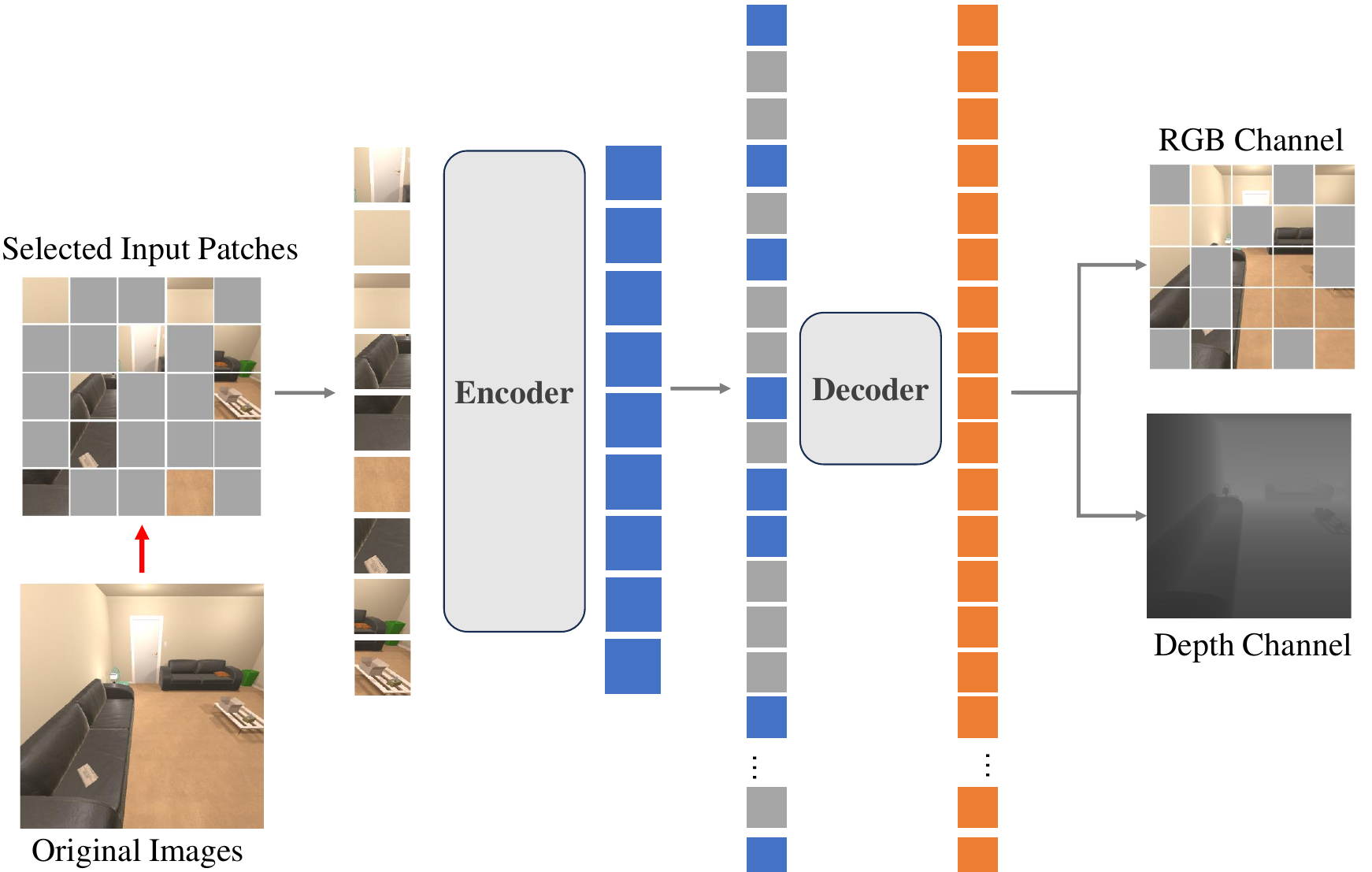}
    \caption{\textbf{Depth Enhanced MAE (DE-MAE)}. During pretraining, a large random subset of RGB image patches is masked out. The encoder is applied to \emph{visible patches}. The full set of encoded patches and mask tokens is processed by a small decoder that reconstructs the original RGB image and also predicts its depth image in pixels using two linear projection heads. After pretraining, the encoder is applied as image feature extractor for navigation tasks.}
    \label{DE_MAE_fig}
\end{figure}
In previous studies, methods for pretraining image feature extractor include supervised learning like classification, or self-supervised learning such as Moco\cite{he2020momentum}, MAE\cite{he2022masked}. Although these pretraining algorithms are widely used, they ignore the particularity of navigation tasks and resulting in the loss of spatial information.
Therefore, We propose depth enhanced MAE (DE-MAE), a self-supervised training algorithm improved upon standard MAE. As shown in  Figure \ref{DE_MAE_fig}, this algorithm requires the decoder to reconstruct not only the masked RGB channels but also the depth channel of the original image.
\paragraph{Encoder and Decoder}
We use the ViT-Base as encoder. The encoder receives an RGB image, and first converts it into patches, then add 2D sine-cosine positional embeddings after the linear projection. After multi-layer transformer operations, the feature map output by the encoder is used as the input of the  decoder.
The decoder employs a significantly smaller ViT network and takes both masked and non-masked patches as input. Finally, two linear projection heads are appended to map the decoder's output tokens to pixel-wise predictions for both RGB and depth channels of the corresponding image patches.
\paragraph{Loss}
Our loss function computes the meansquared error (MSE) between the reconstructed and original images in the pixel space for both RGB and depth channel with identical weights on masked patches (RGB) and all patches (Depth).

\paragraph{Data Collection and Pretraining}
We randomly sampled approximately 2 million RGBD images from training scenes of AI2-Thor. When collecting images, the camera intrinsics keep unchanged, and only horizontal flipping  was employed for data augmentation. In subsequent experiments, we find that using additional augmentation techniques like random cropping will hinder the model's ability to learn the mapping relationship between image and real space, wasting the introduced depth information.
\subsection{Navigation Based On RL}
\subsubsection{Policy Network}
In the image feature extraction, the image feature \begin{math} \boldsymbol{i_t} \end{math} output by ViT pretrained using DE-MAE is obtained. To process the target orientation features obtained from DETR, we use the non-local target memory aggregation (NTWA) module proposed by IOM\cite{IOM}  to obtain the target orientation feature embedding $\boldsymbol{F_t}$. They are fused with previous action embedding \begin{math} PA\end{math}, then
the state representation \begin{math} \boldsymbol{s_t} \end{math} is obtained:
\begin{equation}
  \boldsymbol{s_t}=cat\left( \boldsymbol{\,i_t},  \boldsymbol{F_t} ,\boldsymbol{PA}\right) \boldsymbol{W}
  \label{equ10}
\end{equation}
The LSTM\cite{hochreiter1997long} module is treated as a policy network \begin{math}  \pi \left(\boldsymbol{a_t}|\boldsymbol{s_t} \right) \end{math}, and the asynchronous advantage actor-critic (A3C) algorithm\cite{mnih2016asynchronous} is used to train the  model.
\subsubsection{Reward Function}
In addition to the two-stage reward mentioned above, the comprehensive supervision signal also incorporates rewards and penalties for collisions, navigation termination and other situations. The overall reward structure during reinforcement learning is outlined as follows:

(i) Exploration Reward: As described in Equation \ref{exreward}.

(ii) Distance Reward (penalty) : As described in Equation \ref{navreward}.

(iii) Collision penalty: When an agent collides for the first time, no penalty will be given. If the agent collides again in the same position and orientation, penalty of -0.1 will be given.

(iv) Slack penalty: In each step, a penalty of -0.01 will be given until the episode ends.

(v) Final Reward: A reward of +5 will be given when success.
\section{Experiments}
\subsection{Experimental Setting}
\subsubsection{Dataset}
AI2-Thor and RoboTHOR datasets are selected for evaluation. 
AI2-Thor includes 4 types of room: kitchen, living room, bedroom, and bathroom, each consists of 30 floorplans, of which 20 rooms are used for training, 5 rooms for validation, 5 rooms for testing. 
RoboTHOR consists of 75 scenes, 60 of which are used for training and 15 for validation. 
\subsubsection{Evaluation Metrics}
 Success rate (SR), success weighted by path length (SPL)\cite{anderson2018evaluation} metrics are used to evaluate our method. The formula of SR is \begin{math} SR=\frac{1}{K}\sum\nolimits_{i=1}^K{Suc_i} \end{math}, where \begin{math} K \end{math} is the number of episodes, and \begin{math} Suc_i \end{math} indicates whether the \begin{math} i \end{math}-th episode is successful. 
SPL indicates the efficiency of the agent, its formula is \begin{math} SPL=\frac{1}{K}\sum\nolimits_{i=1}^K{Suc_i}\frac{L_{i}^{*}}{\max \left( L_i,L_{i}^{*} \right)} \end{math}, where \begin{math} L_i \end{math} is the length of the path actually traveled by the agent. 
\begin{math} L_{i}^{*} \end{math} is the optimal path length provided by the simulator.

In the ablation experiment, we uses three additional metrics to evaluate the performance at different stages, two of them are existing metrics Searching Success Rate (SSR)\cite{dang2023multiple} and Navigation Success Weighted by
 Navigation Path Length (NSNPL)\cite{dang2023multiple}.
 SSR is the success rate for the searching stage and is formulated as
\begin{math}
    SSR=\frac{1}{K}\sum_{i=1}^{K} Nav_i
\end{math},  where $Nav_i$ indicates whether the i-th episode enters the pathfinding stage.
NSNPL measures the navigation capability of agents in the pathfinding stage, its formula is \begin{math}
NSNPL=\frac{1}{K_{Nav}}\sum_{i=1}^{K}Suc_i Nav_i\frac{L_i^{*Nav}}{max(L_i^{Nav}, L_i^{*Nav})}
\end{math}, where $K_{Nav}$ is the number of episodes that enters the pathfinding stage. $L_i^{Nav}$ is the path length in the pathfinding stage and $L_i^{*Nav}$ is the  shortest path length in this stage which is calculated according to the starting position of the pathfinding stage.
However, for the searching stage, SSR only considers the results, ignoring the efficiency of the searching and exploring process, and it's easy to achieve full scores for advanced methods\cite{dang2022unbiased,dang2023multiple,IOM}, lacking comparability. Therefore we propose Searching Success Weighted by Searching Path Length (SSSPL)
, a new metric to comprehensively evaluate both the search efficiency.
Its formula is: 
\begin{align}   
SSSPL=\frac{1}{K}\sum\nolimits_{i=1}^K{Nav_i}\frac{L_{i}^{*Search}}{\max \left( L_i^{Search},L_{i}^{*Search} \right)} \end{align} where $L_i^{Search}$ is the actual path length in the searching stage and $L_i^{*Search}$ is the shortest path length in searching stage, it's calculated based on the initial position and all legal positions that satisfies the criterion of the two-stage dividing point.

\subsubsection{Implementation Details}
Our model is trained by 14 workers on 1 RTX 3090 Nvidia GPU. 
We first use DE-MAE in pretraining and then use reinforcement learning to train the agent for 1.5M episodes. 
By evaluating on the validation set, the target confidence threshold $C_{target}$ for dividing point is set to 0.7. To calculate exploration reward, the distances from the agent center to the upper and lower bases of the observable trapezoid region are 1m (0.25m) and 4m (3m) respectively, in a straight-ahead (overhead) view. The learning rate of the optimizer is \begin{math} 10^{-4} \end{math}. $K_e$ and $K_d$ are set to 0.1 and 0.15, and the dropout\cite{srivastava2014dropout} rate of training is 0.45. During testing, the max episode length is set to 100.
We show results for all targets (ALL) and a subset of targets (\begin{math} L\geq 5 \end{math}) whose optimal trajectory length is longer than 5.
\subsection{Comparisons to the State-of-the-art}
\begin{table}
  \caption{Comparisons with the state-of-the-art methods on the AI2-Thor/RoboTHOR datasets}
  \label{table_sota} 
  \begin{tabular}{ c|cc|cc }
    \toprule[1pt]
   \multirow{2}{*}{Method} & \multicolumn{2}{c|}{ALL (\begin{math} \% \end{math})}& \multicolumn{2}{c}{$ L \geq 5 $ (\begin{math} \% \end{math})}\\  
    \cmidrule{2-5} 
    & SR& SPL & SR& SPL \\
    \midrule[1pt]
    SP\cite{yang2018visual}   & 64.70/56.82     & 40.13/29.71    & 54.80/53.72&  37.35/30.95\\
    SAVN\cite{worts} & 65.68/58.14     & 40.60/31.47   & 54.92/54.03&  38.41/30.56\\
    SA\cite{mayo2021visual}   & 68.37/60.49     & 41.04/31.58     & 56.33/55.74&  38.96/31.08\\
    ORG\cite{du2020learning}  & 69.76/62.69     & 39.59/32.35      & 60.78/58.44&  39.58/33.85\\
    HOZ\cite{zhang2021hi}  & 70.65/63.48     & 40.47/32.70     & 62.91/59.46&  40.02/34.42\\
    OMT\cite{fukushima2022object}  & 72.37/66.49  & 33.20/33.76& 61.00/65.85&  30.72/36.55\\ 
    G2SNet\cite{g2s}  & 73.81/64.73     & 41.65/31.82     & 61.69/60.25&  39.21/31.32\\
    VTNet\cite{duvtnet} & 74.73/67.84  & 47.14/34.97& 67.72/66.57&  48.19/36.74 \\
    DOA\cite{dang2022unbiased} &  80.81/71.75  & 46.85/36.58& 74.25/69.30&  48.69/39.15\\
    DAT\cite{dang2022search} & 82.24/73.63  & 48.68/40.66& 75.07/71.35&  49.45/39.89\\
    IOM\cite{IOM} & 83.42/73.95  & 49.38/39.15& 76.10/71.28&  49.94/39.52\\
    MT\cite{dang2023multiple} & 85.47/73.86  & 47.02/38.46&   80.08/71.85&  50.26/38.74\\
   \textbf{Our} & \textbf{86.73}/\textbf{74.93}  & \textbf{50.48}/\textbf{40.37}& \textbf{81.71}/\textbf{73.36}&  \textbf{52.09}/\textbf{39.10}\\ 
  \bottomrule[1pt]
    \end{tabular}
\end{table}
It can be seen from Table \ref{table_sota} that our method surpasses the previous SOTA method (MT)  by 1.26/1.07, 3.46/1.91, 1.63/1.51, 1.83/0.36 (AI2-Thor/RoboTHOR,\%) in SR and SPL. Our method is also superior to the other methods that attempt to implement two-stage strategies only based  on network structure design, such as DAT\cite{dang2022search} and IOM\cite{IOM}.
\subsection{Ablation Experiments}
\begin{figure}
    \centering
    \includegraphics[scale=0.35]{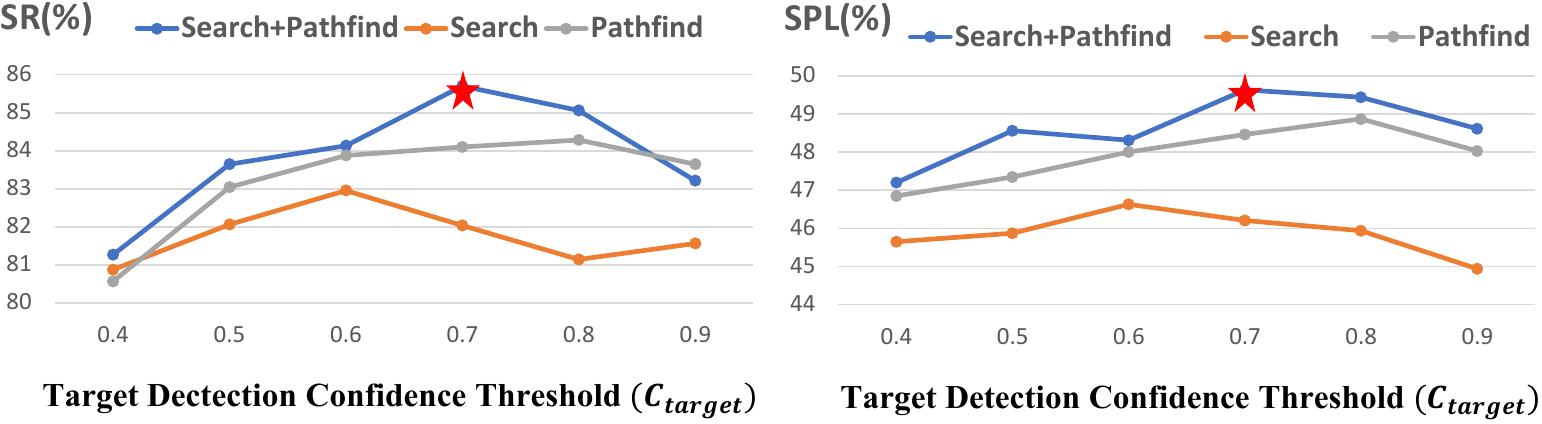}
    \caption{We compared the navigation metrics using different target detection threshold values as two stages' dividing point on AI2-THOR validation set. The red star indicates the choices that optimize the given indicator.}
    \label{line_fig}
\end{figure}
\begin{table*}

        \caption{Ablation experiment results  on AI2-THOR}
  \label{table_abl}
  \resizebox{\linewidth}{!}{  
    \begin{tabular}{ ccc|ccccc|ccccc }
    \toprule[1pt] 
  \multirow{2}{*}{Exploration Reward}& \multirow{2}{*}{Distance Reward}& \multirow{2}{*}{DE-MAE} & \multicolumn{5}{c|}{ALL (\begin{math} \% \end{math})}& \multicolumn{5}{c}{$ L\geq 5 $ (\begin{math} \% \end{math})}\\  
    \cmidrule{4-13} 
    & & &  SR& SPL&SSR& SSSPL & NSNPL & SR& SPL&SSR& SSSPL & NSNPL\\
    \midrule[1pt]
    & &  &                          81.92& 47.15&95.49& 43.87 & 48.06 & 75.76 & 48.37 &93.76& 45.84& 49.72\\
    \midrule[1pt]
    \checkmark & &  &               82.95& 46.63&96.63& 46.06& 48.23 & 77.74 & 48.18 &95.21&47.39 &49.68\\
    & \checkmark &  &               84.28& 48.87&95.98& 45.63& 50.40& 78.93 & 50.74  &94.83&46.95 &52.38\\
    \checkmark & \checkmark & &     85.70& 49.63&97.06& 47.08& 51.56& 80.46 & 51.61 &96.81& 47.92 &53.45\\
     & & \checkmark  &              83.21& 49.35&96.60& 46.20& 50.29& 77.97 & 51.40 &95.12& 47.28 &52.63\\
    \checkmark& \checkmark & \checkmark&\textbf{86.73}& \textbf{50.48}&\textbf{97.79}& \textbf{47.56} & \textbf{52.56}&  \textbf{81.71}&  \textbf{52.09}&\textbf{97.03}& \textbf{49.20} & \textbf{54.10}\\
  \bottomrule[1pt]
    \end{tabular}
}
\end{table*}
To study the effectiveness of the Two-Stage Reward Mechanism and DE-MAE, we conducted ablation experiments on AI2-Thor as shown in Table \ref{table_abl} and  Figure \ref{line_fig}.
\subsubsection{Baseline}
Our baseline model uses the same Resnet18 feature extractor as previous methods\cite{duvtnet,IOM,dang2023multiple}, which is pretrained through the image classification task on ImageNet\cite{deng2009imagenet}. The two-stage reward mechanism is also not applied.

\subsubsection{Two Stage Reinforcement Learning}

As shown in Fig \ref{line_fig}, we tried different target detection confidence threshold value. The excessively  large threshold values  result in  long searching stages, causing the navigation inefficient and prone to missing target. While excessively low threshold may cause the agent to not fully explore the environment and navigate to the wrong target. Therefore, we ultimately set the threshold value $C_{target}$ to 0.7. 

In comparison to the baseline, the incorporation of the exploration reward has led to notable enhancements across multiple metrics, with particularly remarkable advancements in SSR (1.49/1.45, ALL/L$\geq$5, $\%$) and SSSPL (2.16/1.55, ALL/L$\geq$5, $\%$), indicating that the search accuracy and exploration efficiency have been significantly improved. Furthermore, upon integrating distance rewards into the baseline, the NSNPL metric, which serves as an indicator of pathfinding efficiency, has exhibited substantial gains (2.34/2.66, ALL/L$\geq$5, $\%$). When the exploration and distance rewards are combined, the metrics outperform those using either reward individually, demonstrating that rational search and pathfinding behaviors are complementary and cannot be substituted for one another.

\subsubsection{Depth Enhanced MAE}
When DE-MAE is independently integrated into the baseline, most metrics exhibit modest enhancements.
This result suggests that, while the agent's perception capabilities is augmented, it doesn't know how to fully harness the ability for strategy learning. When we combine DE-MAE with TSRM that the improvement of perception serves as the foundation for strategy learning and the pursuit of maximizing exploration and distance rewards, leading to the best outcomes.
\subsection{Irreplaceability of DE-MAE}\label{IRR}
\begin{table}[htb]
    \centering
      \caption{Comparison experiment results for different pretraining method on AI2-Thor}
      \label{DEMAE_tab}
      \resizebox{\linewidth}{!}{ 
 \begin{tabular}{ c|c|cc|cc }
    \toprule[1pt] 
  \multirow{2}{*}{Method(Backbone)} &\multirow{2}{*}{Pretraining Dataset}& \multicolumn{2}{c|}{ALL (\begin{math} \% \end{math})}& \multicolumn{2}{c}{$ L\geq5 $ (\begin{math} \% \end{math})}\\  
    \cmidrule{3-6} 
    && SR& SPL & SR& SPL\\
    \midrule[1pt] 
    Baseline(ResNet18) &ImageNet& 81.92& 47.15 & 75.76 & 48.37 \\
    VC-1\cite{majumdar2024we}(ViT-Base) &Ego4D& 82.12 & 48.19 & 76.46 & 49.76\\
    Standard MAE(Vit-Base) &AI2-THOR&82.76&47.84&76.98&49.61\\
    DE-MAE(ViT-Base) w/ crop  &AI2-THOR& 82.33& 48.36 & 77.05 & 49.91\\
    DE-MAE(ViT-Base) w/o crop &AI2-THOR& \textbf{83.21}& \textbf{49.35} & \textbf{77.97} & \textbf{51.40} \\
  \bottomrule[1pt]
    \end{tabular}}
\end{table}
To demonstrate the irreplaceability of DE-MAE, we introduce other pretraining algorithms and data augmentation approach to the baseline model, the results are presented in Table \ref{DEMAE_tab}.
In baseline, ResNet18 is pretrained by performing image classification task on ImageNet. The VC-1  uses the standard MAE method to pretrain ViT-Base, and its training dataset is Ego4D\cite{grauman2022ego4d} (containing approximately 5.6 million egocentric RGB images). The DE-MAE applies on the same ViT-Base and uses about 2 million RGBD images from training scenes of AI2-Thor as training data. Standard MAE uses the same dataset as DE-MAE without depth channels. We have also tried DE-MAE with random cropping for data augmentation during pretraining.

The experimental results show that using VC-1 improves SR and SPL by $0.2\%$ and $1.04\%$ compared with baseline. This result indicates that the selection of the dataset does have an impact on the pretraining of the agent encoder, the results of Standard MAE approach also verify it.
Though DE-MAE w/ crop uses the depth channel, its metrics has hardly improved compared to Standard MAE.
While the DE-MAE not using cropping has increased the success rate by $0.45\%/0.99\% $ and the SPL by $1.51\%/1.79\%$ (ALL/L$\geq$5).
Since we maintained the same camera intrinsic for navigation agents and pretraining image collection,  random cropping may disrupt this consistency and hinder the model from learning the mapping relationship between images and real space, violating the purpose of introducing depth information.

\section{Conclusion}
In this article, we propose the Two-Stage Reward Mechanism (TSRM) for object navigation and the Depth Enhanced Masked Autoencoder  (DE-MAE) to pretrain the image encoder, enabling the agent to learn effective exploration and pathfinding strategies through sufficient spatial perception capabilities. 
Moreever, to comprehensively evaluate the exploration ability and efficiency of agents, we have proposed a new metric Searching Success Weighted by Searching Path Length (SSSPL).
The comparison experiments demonstrate that our
 work achieves state-of-the-art performance, the ablation experiments and analysis have also demonstrated the effectiveness and irreplaceability of our proposed TSRM and DE-MAE.
Looking ahead, we believe that the sustained research on exploration and pathfinding strategies learning will catalyze further advancements in the field of object navigation.
\section{Acknowledgment}
This work was supported in part by the National Key Research and Development Program of China under Grant No. 2022YFF0712100, the National Natural Science Foundation of China under Grant 62202273, Major Basic Research Program of Shandong Provincial Natural Science Foundation under Grant ZR2022ZD02, Joint Key Funds of National Natural Science Foundation of China under Grant U23A20302, and the Key Technology Research and Industrialization Demonstration Projects of Qingdao under Grant 23-1-2-qljh-8-g


\bibliography{sample}


\end{document}